# Adversarial AI in Insurance: Pervasiveness and Resilience

January 10, 2023

*Elisa Luciano, University of Torino, Collegio Carlo Alberto[1]*

*Matteo Cattaneo, Chief Digital Innovation Officer, Reale Mutua Group[2]*

*Ron S. Kenett, KPA Group, Samuel Neaman Institute, Technion, Israel and University of Torino*

---

[1] *Corresponding Author: elisa.luciano@unito.it*
[2] *The opinions expressed here are those of the Author and do not represent his institution view.*

# Abstract

The rapid and dynamic pace of Artificial Intelligence (AI) and Machine Learning (ML) is revolutionizing the insurance sector. AI offers significant, very much welcome advantages to insurance companies, and is fundamental to their customer-centricity strategy. It also poses challenges, in the project and implementation phase. Among those, we study Adversarial Attacks, which consist of the creation of modified input data to deceive an AI system and produce false outputs. We provide examples of attacks on insurance AI applications, categorize them, and argue on defence methods and precautionary systems, considering that they can involve few-shot and zero-shot multilabelling. A related topic, with growing interest, is the validation and verification of systems incorporating AI and ML components. These topics are discussed in various sections of this paper.

## 1. Introduction

The rapid and dynamic pace of Artificial Intelligence (AI) and Machine Learning (ML) developments is revolutionizing the insurance sector, especially in property and casualty insurance. In 2018, the EIOPA's Big Data Analytics thematic review in motor and health insurance reported that 55% of the firms who participated in the survey were using AI, and 24% were at a "proof of concept" stage. (EIOPA, 2019). EIOPA itself, after noting that data doubles in very short intervals, gives a positive judgment on its use: "Technology is finally making Artificial Intelligence (AI) into a relevant tool to improve our societies. Insurance has been a heavy user of data from practically early days of its existence. The collection of data, even when available, has been expensive. Analysis of this data has been expensive too and often inaccurate. […] The emergence of Big Data (BD) and AI are changing this, making it possible to have more exact knowledge and changing the ways insurers interact with policyholders." (EIOPA, 2021)

The use of AI in insurance helps companies better serve their customers by customizing insurance products and services to fit individual needs, in a more precise and timely manner. AI enters directly as a tool to guarantee the customer centricity of underwriters' actions: an accurate exam of the characteristics, needs and preferences of customers, conducted by filtering their data through AI systems, permits to adhere to them much better than a simplified market analysis did in the past. Services to customers, in all phases of the value chain, can be offered promptly avoiding the mistakes inherent in human actions. Indeed, the advantage of AI does not consist only in the amount of information that it can quickly process, but also in the fact that it can potentially outperform humans in a wide range of tasks, such as content-based processing of information, text analytics, recognition of patterns, trends, and preferences, and improve operational efficiency, as clearly argued in (Eling, Nuessle, and Staubli, (2021). As a consequence, revenues can increase, costs are reduced, and profits and value can go up. It is by now clear that, when AI is appropriately implemented, it has advantages in processing customer information and handling the relationship with clients. AI can handle the enormous amount of information on clients that is collected to exploit predictive analytics.

At the same time, AI generates challenges for underwriters. Unfortunately, the massive use of web-based and computer-based instruments makes every firm and individual, but especially insurance companies, the potential target of cyberattacks. A thorough, clear picture of the risks is in Eling (2020). Among cyberattacks, insurance companies' use of AI can be the target of so-called adversarial AI, which causes AI and ML tools to misinterpret the information provided to them and give output (recommendations, predictions, decisions, categorizations) favourable to the attacker. Information about a category of customers that is collected and processed through an "automatic" AI tool can be distorted by adversarial AI.

We demonstrate that not only a claim model can be defrauded, but predictions based on an attacked system may have better statistical properties and prediction accuracy than the non-attacked ones. High caution and expert judgment based on multiple sources of information are then needed when using AI.

We give general suggestions to prevent attacks and focus on the ability of zero-label countermeasures to prevent them and build resilience against a pervasive phenomenon.

To understand where and how adversarial AI can be exerted, in Section 2 we review some typical uses of AI in property and casualty insurance. In Section 3, we provide a definition of Adversarial AI and some examples of attacks. In Section 4 we categorize attacks based on whether they act on images, audio or text. In Section 5 we discuss how to anticipate and prevent attacks, while in Section 6 we study the issues raised by zero-labelling. Section 7 contains a final discussion.

## 2. How AI is transforming the insurance industry

We distinguish the use of AI in designing new products, from the one in risk assessment, from the substantial help in easing the underwriting process and the whole customer journey, including claim processing. Furthermore, we recall that ML is used for predictions, and is therefore pervasive in the operations of an insurance company.

First of all, AI can be used to improve the design of new products, by tailoring them to the perceived needs of (different categories of) potential customers, differentiated by, say, age, gender, occupation, level of education, experience with the same company, but also by more specific data that they offer through questionnaires or more information that they willingly reveal through their use of social media and friendships. AI enlarges the amount of info that can be used and processed, the stratification of the clients, and the easiness of its processing. In health insurance, for instance, it can give many more details to tailor an insurance product to what a single customer perceives as potential losses relevant for him: medical expenses versus hospitalization periods versus periodic check-ups or follow-up treatments. In car insurance, it can help assess which among insurance against theft, natural catastrophes, vandalism and injuries, are most appreciated by the clients.

Secondly, insurance companies can adopt AI to assess more efficiently and accurately the risk profile of their potential customers, and therefore better tariff their products. In health insurance, for instance, a lot of info beyond past medical history can be processed. Info on the living and nutritional habits, attention to physical and mental wellness and precautionary behaviour of potential customers can be analyze, to better predict the likelihood and amount of claims arriving from him. Also the number of variables that enter the risk assessment of a driver is often decided by an AI algorithm.

Thirdly, AI can improve the underwriting process both ex-ante, by improving the accuracy of the risk evaluation and tariff choice, but also during the underwriting process itself. So, AI can help reduce cost, time and errors in the underwriting processes. As an example, in the past agents had to use their connection to a mainframe to have the data identifying the customer in the company database, and then modify them. With the new technologies, this process is much faster and the signature of a contract takes less time for the customer and the intermediary, with a live execution that helps not only the customer journey but also the intermediaries' and agents' satisfaction with the company.

In general, AI can help interact with the customer before and after the underwriting process, i.e. before and after it becomes a customer. With AI, insurers can use Chatbots and Robo-services to effectively interact, recommend products and handle complaints with customers in a faster, 24/7, and sometimes more personal way. Conversational AI is probably one of the most apparent, cool and engaging applications of AI. Already adopted at a large scale in the banking sector, the possibility of having real-time tailored help through a chatbot is something that most generations like and are by now familiar with, even in the financial domain.

Fourth and importantly, AI can help automate the claims process. This, which is known as E-claim, means that insurance companies provide a (hopefully intuitive) online portal where customers submit claims. The company's management of the claims starts with the submission provided this way instead of by phone or letter. Again, there is an advantage in time, accessibility of the material and clear documentation of the filing.

The simplest application of ML remains the prediction of single event probabilities (frequency) and their severity. In car insurance, for instance, methods for assessing the likelihood of an accident based not only on the drivers' characteristics but also on the road, the weather, the traffic congestion etc. are very useful, especially for short-duration or single-trip insurance policies. These multiple dimensions are used also to assess the likely severity of the damage to the injured cars. ML has proven to be very effective in processing all these data (see for instance Zhang et al., 2018, 2021). Most ML algorithms have been used in this field, from the classical Tree-Based Classifiers or K-nearest neighbours to eXtreme Gradient Boosting (see Gutierrez-Osorio and Pedraza, 2020, Kenett et al 2022a)

## 3. Adversarial Attacks in insurance: definition and some examples

Adversarial AI consists in receiving text or photos or voice messages, depending on the type of AI interaction with the underwriter and its intermediaries (agents, brokers, ...), which cause the underwriter AI and ML tools to misinterpret the information provided to them and give wrong recommendations, or predictions, or decisions, or categorizations. The latter are misleading but favourable to the attacker. Adversarial AI can be used and can damage insurance companies in all AI applications.

When a product is designed to be tailored to the customer's needs, for instance, adversarial AI can be used to intercept the submission of the clients' features and therefore provide a picture of the client that is different from the true one. This means that the product will not be tailored to his true needs anymore.

In the risk assessment process, the info on the customer can be distorted, with the result of misclassifying him and assigning him a risk profile that is lower than the correct one. This means entitling him to a tariff cheaper than due, and creating a potential loss for the company, due to under-reserving. On a large scale, it also means that the company has a wrong measurement of its customer-base quality, has reserves smaller than due and her capital adequacy is ill-based.

As for underwriting, an Adversarial Attack can make it longer and less efficient than forecasted, causing temporary dissatisfaction in the client or the agent. It can also distort the registry of the customer, creating permanent harm to the dataset of the company. Not only: data can be stolen from his registry, breaching his privacy.

During the life of the contract, adversarial AI can interfere with chatbots and conversational AI, for instance by affecting the quality and precision of the answers to the customer, or by distorting his questions to the chatbot. In both cases the result is to lower, sometimes sensibly, the quality of the customer experience, reaching exactly the opposite result of what chatbots are designed for.

Lastly, in claim processing, Adversarial Attacks can be used to magnify the magnitude of the damage suffered by the customer. As a general rule, we know that claim fraud is a pervasive, evergreen problem for insurance companies (see Whitaker, 2019). Making claim filing automatic through AI has obvious advantages in efficiency, but cannot come at the cost of increasing the severity of filing concerning the truth. In a companion paper (Ahmeridad et al., 2023) we provided an example of claim falsification in health insurance, where a small change in the X-ray image was able to make benign cancer classified as malign. By so doing, the attack entitled the patient to a much bigger refund than if the image were not attacked. The appalling feature of such an attack is that the attacked and non-attacked images are indistinguishable by a human eye, and also the statistical properties of the AI algorithm, once attacked, may be better than before the attack, as in the "panda vs gibbon" example below. See also Fynlaison et al. (2019) for applications for health insurance. To have a better understanding of Adversarial Attacks, we provide, in the next section, a characterization for their types.

## 4. Attacks features and consequences

Adversarial Attacks can be first classified based on whether they are image, text or voice-based (see Xu et al., 2020, Zhang and Li, 2018), Zhang et al., 2021).

### 4.1 Adversarial image-based attacks

Several AI applications are image-based. In this case, an Adversarial Attack consists in perturbing the image so little that the perturbed image looks the same as the original to the human eye, but is classified differently (at least with some probability) by the AI algorithm. In many applications based on image classification, including face recognition, researchers use Deep Neural Networks (DNNs) or Convolutional Neural Networks (CNNs) as AI instruments. With the significant improvements in DNN computational power and number of layers, modern DNNs are becoming more and more accurate. The powerfulness of an AI or ML application is usually measured by metrics such as accuracy, which is the number of correct predictions over the total. So, modern DNNs have usually high accuracy both in the train and test set. Nonetheless, they are open to adversarial perturbations, designed to fool the system. Figure 1, for instance, is a famous example of an image that a pre-trained CNN correctly classifies as a "panda" with a probability of 57.7% (left picture). On the right the perturbed image is classified by the CNN as a "gibbon" with 99.3% confidence, even though the output does not look different from the original "panda" image to the human eye (right image). The misclassification obtains by a perturbation which is very small (Goodfellow, Shlens, and Szegedy, 2015). The astonishing fact is that the level of confidence in the erroneous, final classification is much greater than the one of the original image (99.3% instead of 57.7). This happens because the Adversarial Attack is built exactly to maximize the probability of misclassification.

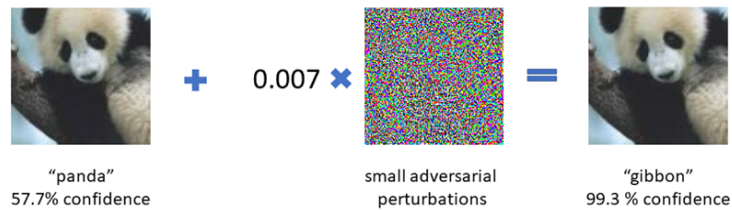

*Figure 1 An Example of Adversarial Perturbations and Adversarial Example (Goodfellow et al, 2015))*

Telemedicine can also be affected by adversarial AI. A wide range of image processing studies show that AI helps to diagnosis illnesses very accurately, making telemedicine an attractive service for modern insurance companies acting in the health sector. However, any effort can be vanished if the corresponding images are not trustworthy (see Wetstein et al., 2020).

Another image recognition technique in insurance that may be threatened by Adversarial Attacks is the process of early warning and damage detection in natural catastrophes. For example, in extreme events such as tornadoes, hurricanes, and wildfires, image analysis from satellite pictures can be used to provide early warning systems (e.g., Swiss Re), especially when a parametric insurance is active. However, if the image is attacked so as to hide the early warning need, the catastrophe can explode in its full power, with no prevention in place. Similarly, trained image recognition models are sometimes applied to determine at least by first approximation the level of damage in areas hit by a tornado or a flood, where parametric insurance is active. The application of an AI image-based algorithm subject to even a slight perturbation through an Adversarial Attack can cause a major underestimate of the loss following the natural catastrophe.

As for car insurance, falsified pictures of the damage, only slightly different from the truth, as in the panda/gibbon case, can make the insurance company open a reimbursement procedure for much bigger damage than the real one. In the car case, attacks may hit not only the insurance company, but also the car producer. Think of self-driving cars, or cars with sophisticated sensors and automatic reactions. Adversaries can attack an autonomous vehicle by changing the environment so that the vehicle's camera systems are no longer able to correctly process image data (Sarker et. al., 2020). In (Evtimov et al., 2017), the authors showed that graffiti on a 'Stop' sign can fool a DNN into recognizing it as a 'Speed Limit 45' sign, and this could lead to a self-driving car speeding up at a stop sign. In these cases, the efforts of the regulators of car producers to avoid fooling the autonomous car self-driving capabilities go hand in hand with the interests not only of passengers, but also of insurance companies.

### 4.2. Adversarial text-based attacks

Another category of Adversarial Attacks is the one that aim at voice recording or written text. Voice recording means continuous wavelets. These can be attacked more or less like images. Figure 2 for instance shows that an imperceptible, malicious interference on the top wavelet, which is automatically transformed into text by AI and whose meaning is written on the right of the figure, transforms it into a sentence which is interpreted by AI in a completely different way (bottom wavelet, meaning on the right). This example is from Carlini and Wagner (2018), who, to raise awareness of Adversarial Attacks in IoT, created attacked versions of automatic speech-to-text transformation systems as used in "Apple Siri" and "Google Now". In their speech recognition example, a slight bias — which is almost inaudible — creates strong, iterative, optimization-based attacks that transform any audio waveform into any target transcription, with a very high success rate.

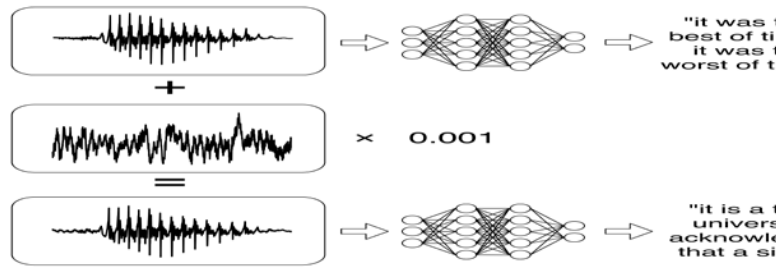

*Figure 2 Adversarial Audio- any Audio Waveform turns Into any Target Transcription with High Rates Success by Only Adding a Slight Distortion, Carlini and Wagner (2018).*

Most of the Robo-Advisors or chatbots can handle complaints and communication with customers (Eling et al., 2021) by receiving written audio waveform, and then converting it into computer-readable structured by ML techniques. This exposes them, following Carlini and Wagner's example, to serious risks.

While image and audio data are continuous, i.e., consist of pixels or audio frequencies with a continuous distribution of values, other complex data, such as text, is discrete and does not consist of quantifiable values. It can be more challenging to create adversarial examples for discrete structures, such as text, because the semantic and syntactic properties of the original input, which are perceived by humans, should be preserved by the attacked text (Lei et al., 2018). So, the relationships with customers, including Robo-Advice and chatbot, exclusively based on text, happen to be more resilient. The great benefits that most insurance companies, including both the native digital ones, like Lemonade, and the traditional ones receive from the application of text analytics and written language processing, including sentiment detection, for pricing and customization of insurance products, seem to be quite sheltered, for the time being, from attacks.

## 5. Anticipating Adversarial Attacks

In a survey recently conducted by EIOPA (see EIOPA, 2021) cyber risks, which include adversarial AI risks, were listed fifth among the exceptionally important risks. They were preceded by "meeting regulatory requirements", "trust from consumers", "data accuracy issues" and "fairness considerations", and followed by several other risks, including "reputational risks " and "project risks", which included the risk of misinterpreting the consumer needs. In light of the explanations given above, we claim that Adversarial Risks arise as cyberattacks, but may entail significant reputational and project risks, to mention a few (see also Eiling, 2020).

At the same time, the illustration above should have given a measure of how differentiated attacks can be, both in terms of field of application, aim and AI method (image versus audio or text) attacked. A universal, bullet-proof patch is far from existing, while the list of specific countermeasures is rich and well-motivated (see for instance Papernot, 2018, Qiu et al., 2019, Ren et al., 2020, Warr, 2019)

Insurance companies are not alone in willing to prevent specific attacks. We have seen above that resilience to attacks on self-driving cars, implemented by producers themselves, for instance, contributes to preventing car accidents. Therefore, they can cooperate in prevention with members of enlarged service ecosystems.

Let alone specific countermeasures and cooperation with larger ecosystems in prevention, the general suggestion we can give is the usual one: when considering AI adoption, do multiple tests and be cautious. Consider AI and its opportunities, but also the risks we put into evidence here, as part of the bigger data quality and usage policy of the company, putting pros and cons into perspective, being keen on maintenance and experts' opinion.

However, among the many prevention mechanisms that can be put in place in an insurance company, there is one that is worth emphasizing here: the zero-label or few-shots nature of Adversarial Attacks. This means that often Adversarial Attacks generate a new category among multi-labels, that either had not been seen before (zero-label), or had rarely been seen before (few-shots label). Therefore, particular attention to the treatment of zero and few labels is also a way to prevent cyberattacks, seldom noticed in the literature. The next Section explains this particular prevention measure.

## 6. Preventing Adversarial AI through zero and few-shot label accurate treatment.

Fortunately, Adversarial Attacks are, most of the time, events never seen before. Unfortunately, because of that, they can be classified as zero-shot classification events. Zero-shot classification, in ML, means that we train a model classifying objects into some classes and predict a brand new class, which the model has not encountered in the training data, the attacked one (Rios and Kavuluru, 2018, Huang. 2020)

The first thing that comes to mind is to include the class name in the list of classes, even though there are no training samples for this class, and provide a general, high-level description of the class, putting it into relation with classes that the ML algorithm has been trained to. As usual, this happens in the training phase. The inference phase must be designed accordingly. However, how does one implement the idea and what is the algorithm's reliability in the presence of these zero-labels?

A very simple approach has been suggested by Romera-Paredes and Torr (2015). They consider explicitly the training-inference steps in a linear model that describes the links between features, attributes and classes. Linear models encompass several categories of ML. The first step, in the training phase, learns the relationship between attributes and features. The second step works in the prediction phase and considers the links between attributes and classes. Formally, the Authors do not minimise – through an appropriate choice of parameters or weights  - the loss coming from the weighted product of features over the sample, because here the attributes of the class never seen before are not used. Instead, they collect the attributes of the classes into a new matrix, and use a product containing this matrix to perform the minimization in the training phase. By so doing, they include the attributes of the class never seen before. In Romera-Paredes and Torr, the approach is applied to the classical animal-image processing problem that we reported also in the figures above. Their model is asked to be able to categorize an animal present in the validation – but not the training - data through the insertion of its attributes in the training phase and detection of them in the inference phase. Though so simple, the approach is shown by the Authors to outperform the previously-suggested approaches by 17%.

A more sophisticated approach stays in Hu et al. (2020), who address the problem of misdiagnosis in industrial or health applications, a problem that we have seen to be relevant for insurance applications. They use a data-augmentation algorithm, namely resampling, and a CNN. They show on two datasets that their approach is robust and efficient.

Another way of addressing the problem is in Yang et al. (2022), who, concerning industrial applications, propose a sophisticated called a multi-source transfer learning network (MSTLN), to transfer knowledge from several datasets – experiments on existing machines, each of one of which might present a rarely seen label – and process them through a single diagnostic module.

Domain adaptation techniques that handle rare events zero shot situations can be categorized by two properties: what the learner tries to learn and how the learner learns it. What the algorithm tries to learn can be divided into three categories: (1) learning invariant function between the source and the target domain, where the function on one hand uses invariant features of the source and the target domain and on the other hand successfully classify data in the source domain. (2) learning two different functions for the source and the target domains where some of their properties are similar or identical, for example, using the same features extractor but a different classifier. (3) Learning a mapping function that maps examples from the target domain to the source domain (or the other way) and learning a function that successfully classifies data in the source domain (or the target domain, respectively).

How the learner learns can be categorized into four categories: (a) minimizing distribution metrics between the extracted features of the source and the target domain like maximum mean discrepancy (MMD), coral and central moment discrepancy (CMD) etc. (b) Using adversarial approaches. For example, a feature extractor E learns to extract features from the source and the target domains, and a discriminator D competes with E for leaning to find if the extracted features correspond to an example from the source or the target domains. Parallel to that, another network learns to classify the extracted features based on the labelled examples of the source domain. (c) Batch-normalization methods. (d) parameters transfer methods where the network is first pre-trained using the source domain and then tuned with examples from the target domain.

In the context of mechanical systems, Leturiondo et al. (2015) refer to the case of zero-fault shot learning (labelled and unlabeled) and suggest using simulated data as the source domain to classify unseen faults in the target machine. In this context, Sobie et al. (2018) also refer to the case of zero-fault shot learning (labelled and unlabeled) and suggest using simulated data as the source domain to classify unseen faults in the target machine. They showed and emphasized that simulated data can help in the diagnosis as a first stage but when real data is added the diagnosis results are much more accurate. Their study show how simulation with a preprocessing of signals can achieve zero-fault shot learning.

## 7. Discussion

The paper covers a range of topics related to AI and ML. In particular, it expands on the implications of Adversarial AI to insurance companies and discusses zero-shot methods that apply when systems are designed to handle rare events. It also refers to applications to mechanical systems where lessons learned can be relevant to insurance applications.

The approach by Romera-Paredes and Torr (2015), to Adversarial Attacks, can be considered an omnibus procedure with low implementation costs. As such, it could be an omnibus way of trying to prevent Adversarial Attacks too. The last two models discussed here are more appropriate when images and machines – such as X-rays or laboratory ones in health – are considered. Since health refunding can be costly for insurance companies, it might be worth considering them in insurance applications.

The topics of Adversarial Attacks and Zero-Shot labelling pose significant challenges to the validation and verification of systems used in insurance systems and in general (Pullum, 2021). An example showing the importance of proper validation procedures in predictive analytics is presented in Kenett et al (2022b). The method described there, befitting cross-validation (BCV), is based on an assessment of the data generation process. In the context of personal insurance, the use of the individual as a unit for stratification, as opposed to a transaction, can strongly affect the stratification of data into training and validation.

The topics discussed in the paper require proper preparation and resilience building by insurance companies in domains enveloping the development and deployment of AI and ML, given the pervasiveness of adversarial AI.